%% file: paper.tex
\documentclass[10pt, a4paper]{article}

\usepackage{lrec-coling2024}
\usepackage{amsmath}
\usepackage{listings}
\usepackage{tikz}
\usepackage{lipsum,adjustbox}
\usetikzlibrary{shapes, arrows, positioning}
\usepackage{graphicx}
\usepackage{tipa}

\newcommand{\cat}{\textrm{\normalfont \textctc}}
\newcommand{\um}{\textrm{\normalfont \textltailm}}
\newcommand{\tm}{\textrm{\normalfont \textturnt}}
\newcommand{\sem}{\textrm{\normalfont \textesh}}

\title{Low-Resource Machine Translation through Retrieval-Augmented LLM Prompting: A Study on the Mambai Language}
\name{Raphaël Merx$^\cat$ ~~~~~~  Aso Mahmudi$^\um$~~~~~~ Katrina Langford$^\tm$
 \\ {\bf \large Leo Alberto de Araujo$^\sem$ ~~~ Ekaterina Vylomova$^\um$}}

\address{$^\cat$Catalpa International~~~ $^\um$The University of Melbourne ~~~$^\tm$Timorlink ~~~$^\sem$Seminario Menor Balide Dili \\
         \texttt{raphael.merx@gmail.com} ~
         \texttt{timorlink@hotmail.com} ~
         \texttt{amahmudi@student.unimelb.edu.au} ~\\
         \texttt{leonberto372@gmail.com} ~
         \texttt{vylomovae@unimelb.edu.au}~
    }

\abstract{
This study explores the use of large language models (LLMs) for translating English into Mambai, a low-resource Austronesian language spoken in Timor-Leste, with approximately 200,000 native speakers. Leveraging a novel corpus derived from a Mambai language manual and additional sentences translated by a native speaker, we examine the efficacy of few-shot LLM prompting for machine translation (MT) in this low-resource context. Our methodology involves the strategic selection of parallel sentences and dictionary entries for prompting, aiming to enhance translation accuracy, using open-source and proprietary LLMs (LlaMa 2 70b, Mixtral 8x7B, GPT-4). We find that including dictionary entries in prompts and a mix of sentences retrieved through TF-IDF and semantic embeddings significantly improves translation quality. However, our findings reveal stark disparities in translation performance across test sets, with BLEU scores reaching as high as 21.2 on materials from the language manual, in contrast to a maximum of 4.4 on a test set provided by a native speaker. These results underscore the importance of diverse and representative corpora in assessing MT for low-resource languages. Our research provides insights into few-shot LLM prompting for low-resource MT, and makes available an initial corpus for the Mambai language.
 \\ \newline \Keywords{low-resource languages, austronesian language, large language models, prompting, dictionary, parallel data} }

\begin{document}

\maketitleabstract

\section{Introduction}


Large language models (LLM) have shown remarkable abilities to perform natural language processing (NLP) tasks they were not explicitly trained for, including named entity recognition \cite{llm-ner}, text classification \cite{llm-classification}, text summarisation \cite{llm-summarisation}, and machine translation \cite[MT]{hendy2023good,kocmi-etal-2023-findings,chowdhery2022palm}. LLMs can be competitive with traditional encoder-decoder MT models for high-resource languages, but lag behind traditional MT models when translating to and from low-resource languages \cite{robinson-etal-2023-chatgpt,hendy2023good,garcia2023unreasonable}. 

While LLMs can achieve moderately high translation accuracy through zero-shot prompting \cite{wang2021language}, few-shot prompting can improve translation accuracy \cite{pmlr-v202-zhang23m}. Research on the selection of example sentences for use in LLM prompts found that examples close to the source text do not always result in better translation than random examples \cite{vilar-etal-2023-prompting}, but that in-domain examples can improve accuracy for technical domains \cite{agrawal-etal-2023-context}. In particular, for English to Kinyarwanda MT, \citet{moslem2023adaptive} finds an improvement of 11 ChrF points when using in-domain examples instead of random ones.

Using domain adaptation as an analogy, in this paper we explore whether LLMs can be prompted to translate \emph{into} a very low-resource language, through careful selection of sentences and words close to the source text for use in prompting. We work with the Mambai language, a primarily oral language from Timor-Leste with around 200,000 native speakers \cite{Census2015}. We source prompt examples exclusively from \citetlanguageresource{hull2001mambai}, a language manual which includes parallel English-Mambai sentences and a bilingual word dictionary. We evaluate machine translation quality on both a random subset of sentences from the manual, and on a small corpus of translations collected from a native Mambai speaker.

We find that translation accuracy varies a lot depending on (1) the test set used for evaluation, (2) LLM used for translation, and (3) examples included in the prompt. While 10-shot translation yields BLEU score as high as 23.5 for the test sentences sampled from the language manual used in prompting (with GPT-4 and a mix of sentences retrieved through semantic embeddings and TF-IDF in the prompt), BLEU drops below 5 across all experimental setups for test sentences outside of this domain (novel sentences collected from a native speaker).\footnote{We release the code for extracting the language manual data and for using this data to construct a few-shot prompt given a sentence to translate, as well as the corpus of sentences translated by the paper's author, in \href{https://github.com/raphaelmerx/mambai}{https://github.com/raphaelmerx/mambai}. The language manual data is available upon request.}

Our findings highlight the risks of relying on a single source when evaluating MT for low-resource languages, especially for languages like Mambai that do not have a standardised vocabulary, orthography, or syntax, where a single corpus can have substantial influence on NLP experiments, despite not always being representative of the language's variations.

\section{The Mambai Language}
\label{sec:background-mambai}

Timor-Leste (also known as East Timor) is a half-island nation in South-East Asia, with a population of 1.3 million as of 2022 \cite{Census2022}. While its official languages are Portuguese and Tetun Dili  \cite[also spelled Tetum]{Constitution}, the country has over 30 indigenous languages, from both the Austronesian and Papuan language families \cite{kingsbury2010national}.

Mambai (also spelled Mambae) is the country's second most common mother tongue after Tetun, with around 200,000 native speakers \cite{Census2015}. An Austronesian language, it is mostly spoken in the Ermera, Aileu, Manufahi, and Ainaro municipalities \cite{timorlanguages}, and does not have a standardised orthography \citeplanguageresource{hull2001mambai}. It has three distinct varieties, and this article will focus on the southern variety, spoken primarily in the Ainaro, Same, and Hatu-Builico administrative posts \cite{fogacca2013estudo}.

Translating to Mambai can bring valuable material closer to Mambai-speaking communities. For example, the Government of Timor-Leste has a mother tongue education program named EMULI, which found that students who were taught in their mother tongue have a higher level in reading comprehension and mathematics than students taught in Portuguese. This program leverages translated material for the curriculum \cite{kirstyswordgusmaoQualityInclusive, embli-eval}. 

Unfortunately, in the taxonomy of \citet{joshi-etal-2020-state}, Mambai would be assigned class 0, "The Left-Behinds", i.e. ``languages that have been and are still ignored in the aspect of language technologies''. A search for Mambai sentences on OPUS \cite{opus} returns only 36 sentences, all from Tatoeba.\footnote{\href{https://tatoeba.org/}{https://tatoeba.org/}} To our knowledge, the only NLP tools that claim to support Mambai are language identification models GlotLID \cite{glotid} and MMS \cite{mms}. Mambai does not appear on  popular datasets for low-resource languages such as MT560 \citeplanguageresource{gowda-etal-2021-many} or FLORES-200 evaluation benchmark \citeplanguageresource{flores2022}.

\begin{figure*}[!ht]
\centering{
\resizebox{\textwidth}{!}{\input{data_extraction_process}}}
\caption{Overview of our process for extracting dictionaries and a parallel corpus from the Mambai Language Manual} \label{fig:process-data-extraction}
\end{figure*}
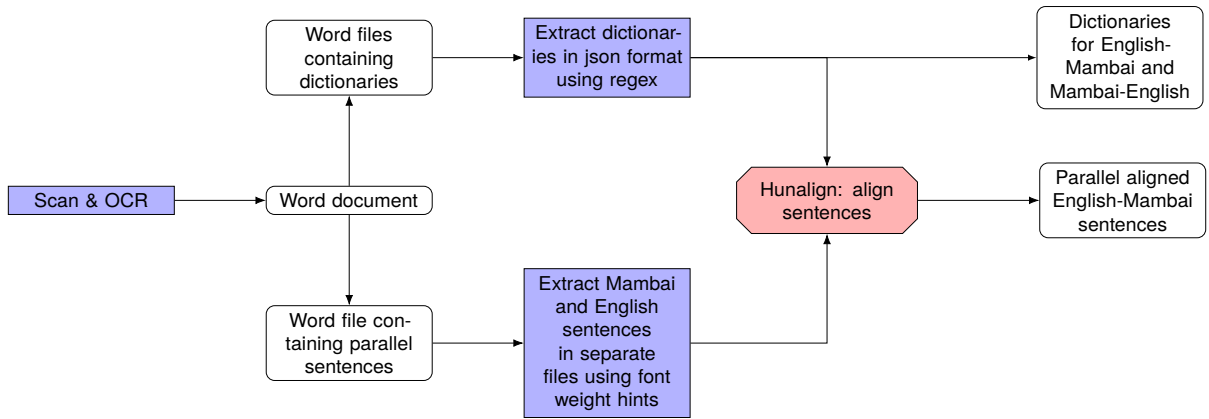

\section{Methodology for Data Extraction}
As the language does not have any resources in a machine-readable format, we start by digitising the available materials. The general process of data extraction is illustrated in Figure~\ref{fig:process-data-extraction}.

\begin{figure}[!ht]
\begin{center}
\includegraphics[scale=0.5]{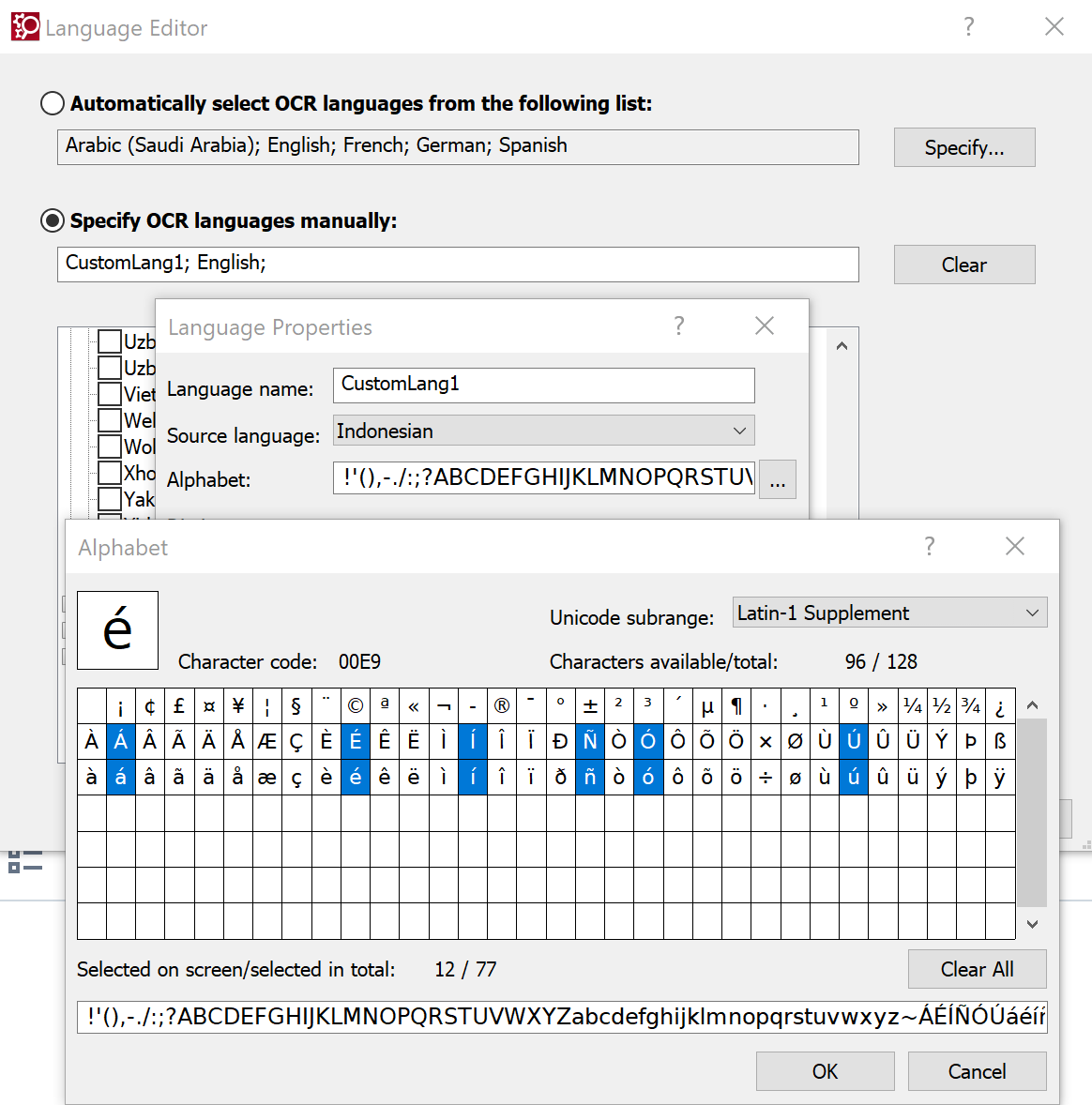} 
\caption{Mambai configuration in ABBYY FineReader 15.}
\label{abbymambai}
\end{center}
\end{figure}

\subsection{Materials}

Our primary data source is a Mambai Language Manual \citelanguageresource{hull2001mambai} that aims to teach the basics of Mambai to foreign speakers, following the Ainaro variety. This 109-page long document includes a pronunciation guide, a grammar, a phrase book, and bilingual dictionaries (English-Mambai and Mambai-English).\footnote{The author of this book gave his consent to us using it as material, and we acknowledge him as the holder of copyright protecting this intellectual property.}

To test generalisation of our results, we collaborated with a native Mambai speaker who translated a small corpus of 50 English sentences to Mambai. Since Mambai has no formalised orthography, we tried to keep  orthography close to that used in the manual, however we did not aim to produce the same syntactic structures as the manual.

\subsection{OCR Process}
\label{sec:ocr}

For the Mambai Language Manual, which we received in paper format, we followed the following OCR process:
\begin{enumerate}
    \item The book was scanned using an optical zoom camera, which reduces the radial distortion effect and improves the OCR quality;
    \item The open-source ScanTailor software\footnote{\href{https://scantailor.org/}{https://scantailor.org/}} was employed to semi-automatically deskew images and make them flat black and white;
    \item In the proprietary software ABBYY FineReader 15,\footnote{\href{https://pdf.abbyy.com/}{https://pdf.abbyy.com/}} we set up a language alphabet, taking into account the characters utilised in each book, with Indonesian (also an Austronesian language) serving as the fallback language, as illustrated on Figure~\ref{abbymambai}. The result of the OCR process was saved in a Word document, preserving font formatting;
    \item We then manually separated the extracted data into three collections:
    \begin{enumerate}
        \item the section of the manual that contains parallel sentences (14,347 words),
        \item the section that contains the English to Mambai dictionary (4,023 words),
        \item the section of the manual that contains the Mambai to English word dictionary (4,522 words).
    \end{enumerate}
\end{enumerate}

\subsection{Text Corpora}

In this subsection, we present the process of our corpus construction: using the Word documents produced in Section~\ref{sec:ocr}, we create English-Mambai bilingual dictionaries in JSON format and a corpus of parallel English-Mambai sentences.

\subsubsection{Dictionary extraction}
\label{sec:dictionary-extraction}

For dictionary files, we mined triplets (entry, part of speech, translation) through the following process:
\begin{itemize}
    \item using the python-docx library,\footnote{\href{https://python-docx.readthedocs.io/}{https://python-docx.readthedocs.io/}} read the file by preserving font weight, and identify text in bold as the dictionary entry;
    \item use a regular expression to match the part of speech, if any;
    \item use the rest of the text as value corresponding to the entry;
    \item if one entry had multiple translations, denormalise them by splitting with ``;'' and ``,''.
\end{itemize}

This process outputs dictionaries in JSON format, one for the English to Mambai direction (1,790 entries), and one for the Mambai to English direction (1,592 entries). Where present, each entry also contains part of speech information, e.g.

\begin{verbatim}
{
  'entry': 'beik',
  'translation': 'silly',
  'part_of_speech': 'adj.'
}
\end{verbatim}

\subsubsection{Parallel sentence extraction}
\label{sec:parallel-sentence-extraction}

Since no embedding models or MT systems support Mambai, we were precluded from relying on sentence embeddings \cite{Thompson2019} or back-translations \cite{sennrich-volk-2011-iterative} to mine parallel sentences from extracted documents. Instead, we rely on a combination of Gale-Church sentence-length information \cite{gale-church-1993-program} and lexical similarity through the Hunalign\footnote{\href{https://github.com/danielvarga/hunalign}{https://github.com/danielvarga/hunalign}} sentence aligner \cite{Varga2007}.

We identify Mambai sentences from their bold font-weight, English sentences from their normal font-weight, and section delimiters through text in upper case. For each section, we put the set of Mambai and English sentences in separate text files, which are fed to Hunalign, along with the bilingual dictionary extracted in Section~\ref{sec:dictionary-extraction}. Hunalign outputs a series of tab-delimited aligned sentence pairs, with an alignment score for each pair. After manual review of a subset of 100 sentences, we find that setting a score threshold of 0.2 corresponds to keeping a high number of well-aligned sentences, while removing poorly aligned ones. After filtering out sentence pairs below this threshold, we land on 1,187 parallel sentences extracted from this phrase book, from a total of 1,275 potential bitexts.

Since sentences come from a language education manual, they tend to be relatively short, with an average of 5.05 words per sentence in Mambai, and 5.66 words per sentence in English. Some sentences have alternative words in parentheses, which we leave in place, e.g.:

\begin{verbatim}
{
 'Mambai': 
  "Baléb pôs masmidar lao xa (kafé).",
 'English': 
  "Don't put sugar in my tea (coffee)."
}
\end{verbatim}

\section{Mambai Translation through Retrieval-Augmented LLM Prompting}
\begin{figure*}[htb]
\centering{
\resizebox{\textwidth}{!}{\input{translation_process}}}
\caption{Overview of our process for translating English sentences to Mambai using both dictionary entries and sentence pairs in few-shot LLM prompting.} 
\label{fig:process-translation}
\end{figure*}
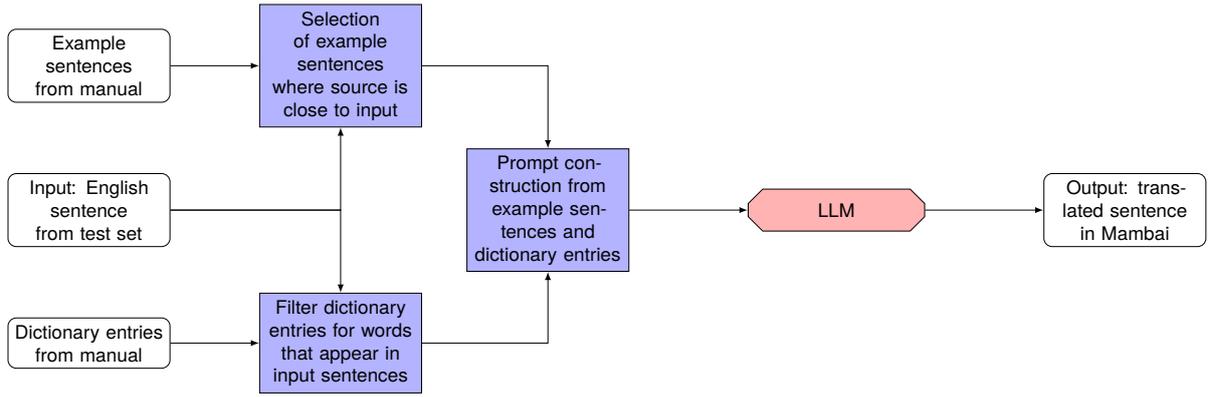

After all required data is ready, we now turn to the machine translation part.
The general process for translation is illustrated in Figure~\ref{fig:process-translation}.

\subsection{Rationale}

\citet{adelani-etal-2022-thousand} found that a couple thousand high-quality sentences can substantially increase low-resource MT performance, giving us hope that a language manual with a similar order of magnitude of data could be enough to produce moderate-quality translations.

Working with LLM prompting gives us a flexible format to incorporate both the parallel sentence corpus and the dictionary entries. Further, having access to a phrase book offers substantial domain coverage, in comparison with corpora purely from the religious domain, which are often the only option for low-resource languages \cite{haddow-etal-2022-survey, embli-eval}.

Here we work on English to Mambai translation, aiming to address the following research questions:
\begin{itemize}
    \item Given an English sentence, how can a corpus of bilingual sentences, and a bilingual word dictionary, be incorporated in an LLM prompt to maximise translation accuracy?
    \item Which LLMs (open-source or proprietary) show the best results for translating into a low-resource language, and what is the observed variance between them?
    \item How does translation accuracy vary across test sets?
\end{itemize}

\subsection{Methodology}

\subsubsection{Data setup}

Our bilingual corpus of 1,187 parallel Mambai-English sentences is randomly split into 119 (10\%) sentences used for testing translation, and 1,068 (90\%) sentences for potential use in the prompt, after retrieval selection. Since our objective is to translate full sentences, not individual words, all 1,790 words in the Mambai dictionary are used in prompting.

We also assess translation system quality by providing a different test corpus of 50 sentences translated from English to Mambai by a native speaker of Mambai. This small corpus has relatively simple but slightly longer sentences, with 9 words per sentence on average. The English source sentences were designed to cover a broad range of domains, such as daily life activities, education, health and well-being, family relationships, religion, politics, weather, employment, food and agriculture, technology, personal characteristics, and Timor-Leste specific historical events.

By using the two test sets, we aim to evaluate robustness to variance between domains, as well as estimate risks of overfitting that come from using a test corpus that comes from the same material as the data for prompting. Expected variance between test sets comes from their different authors, their different years of publication (2001 vs 2024), and potentially by them covering different domains.

\subsubsection{Prompt}

We make use of the best performing prompt template from \citet{peng-etal-2023-towards}, to which we add dictionary entries for words found in the sentence, landing on the following prompt template:

\lstset{
  basicstyle=\ttfamily,
  columns=fullflexible,
  breaklines=true,
}

\begin{lstlisting}
You are a translator for the Mambai language, originally from Timor-Leste.

# Example sentences

English: {Sent_eng_1}
Mambai: {Sent_mgm_1}

English: ...
Mambai: ...

# Dictionary entries

English: {Word_eng_1}
Mambai: {Word_mgm_1}

English: ...
Mambai: ...

Please provide the translation for the following sentence. Do not provide any explanations or text apart from the translation.

English: {input}
Mambai:
\end{lstlisting}

\begin{table*}[!ht]
\begin{tabularx}{\textwidth}{Xllllll}
\hline
\textbf{Model} & \textbf{\(N_{\text{TFIDF}}\)} & \textbf{\(N_{\text{embed}}\)} & \textbf{\texttt{UseDict}} & \textbf{BLEU} & \textbf{ChrF} & \textbf{ChrF++} \\ \hline
gpt-4-turbo & 0 & 0 & FALSE & 3.7 & 22.4 & 19.9 \\
gpt-4-turbo & 0 & 0 & TRUE & 6.9 & 25.3 & 24.7 \\
gpt-4-turbo & 10 & 0 & FALSE & 16.1 & 40.3 & 39.7 \\
gpt-4-turbo & 10 & 0 & TRUE & 20.9 & 41.8 & 41.6 \\
gpt-4-turbo & 0 & 10 & FALSE & 16.8 & 38.2 & 37.4 \\
gpt-4-turbo & 0 & 10 & TRUE & 18.3 & 39.6 & 39.5 \\
gpt-4-turbo & 5 & 5 & FALSE & 17.7 & 40.4 & 39.6 \\
gpt-4-turbo & 5 & 5 & TRUE & \textbf{21.2} & \textbf{41.8} & \textbf{41.6} \\
Mixtral 8x7B & 5 & 5 & TRUE & 9.0 & 30.9 & 30.4 \\
LlaMa 70b & 5 & 5 & TRUE & 12.3 & 32.3 & 31.8 \\
\end{tabularx}
\caption{Experiment results for test set from the language manual. \(N_{\text{TFIDF}}\) and \(N_{\text{embed}}\) represent the number of sentence pairs retrieved through TF-IDF and semantic embeddings, respectively. \texttt{UseDict} indicates whether dictionary entries are included in the prompt. While different hyperparameter combinations were tested for all models, we only report on the best configuration for the less performant models (Mistral 8x7B and LlaMa 70b).}
\label{tab:experiment_results_manual}
\end{table*}
\subsubsection{Models}

We experiment with three models: \textbf{Mixtral} as it is the open-source model with the highest MT-bench score \cite{jiang2024mixtral}, \textbf{LlaMa 70b} \cite{touvron2023llama} as it has a permissive license and has shown high zero-shot translation performance \cite{xu2024a}, and \textbf{GPT-4}, which, despite being proprietary, has very high zero-shot translation performance \cite{xu2024a}.

For each model, we experiment with the following setups:
\begin{itemize}
    \item \texttt{UseDict} (either \texttt{True} or \texttt{False}): For each word that appears in the source language input (English), if this word is present in the English-Mambai dictionary, we include its dictionary translation in the prompt;
    \item \(N_{\text{TFIDF}}\): Number of sentence pairs retrieved through TF-IDF, where the English sentences are ranked according to TF-IDF similarity to the input. The rationale here is that less frequent words can be harder to translate, therefore should be surfaced in the prompt more often. \(N_{\text{TFIDF}} \in \{0, 5, 10\}\)
    \item \(N_{\text{embed}}\): Number of sentence pairs retrieved through LASER semantic embeddings \cite{touvron2023llama}, where the English sentences in training set are first ranked using cosine similarity to the input. \(N_{\text{embed}} \in \{0, 5, 10\}\), similar to \citealp{pmlr-v202-zhang23m, vilar-etal-2023-prompting, hendy2023good}.
\end{itemize}

For each combination of the above features, we measure the BLEU and Chrf++ scores on both test sets, one from the language manual, and one manually translated by a native speaker.

\subsection{Translation Results}

Our experiment results for test sentences from the manual are provided in Table~\ref{tab:experiment_results_manual}, and Table~\ref{tab:experiment_results_leo} provides the results for the test set collected from a native speaker.

To summarise, we make the following observations:

\textbf{(1) Translation accuracy varies widely between both test sets}. While we get an accuracy of up to 23.5 BLEU (41.9 ChrF++) for the test set that comes from the language manual, we could not reach a BLEU higher than 4.4 (33.1 ChrF++) for the test set from the native speaker. More analysis is needed to understand this discrepancy, but it sends a strong signal about the risks of overfitting by using a test set that comes from the same material as the examples used in prompting. In particular, we think our result might partially invalidate \cite{tanzer2024mtob}, which similarly attempts to translate into a very low-resource language using prompting from a single grammar book, but used exclusively sentences from the grammar book in the test set.

\textbf{(2) Dictionary entries help improve translation quality}. When including dictionary entries in the prompt, filtering on words that appear in the source text, we found that translation quality improved significantly. This is true across all experiments when keeping other hyperparameters constant, with an average improvement of 3.25 BLEU points and 2.7 ChrF++ points.

\textbf{(3) A blend of sentences retrieved through semantic embeddings and through TF-IDF yields the highest translation accuracy}. When working with a random split of sentences from the language manual in particular, a blend of 5 sentences retrieved through TF-IDF and 5 sentences retrieved through semantic embeddings outperforms 10 sentences retrieved exclusively through one of these features. This holds true for all three LLMs tested in this project.

\textbf{(4) GPT-4 consistently outperforms other LLMs}. GPT-4 yields both the highest translation score overall, and the higher translation score for every single experiment, when compared with LlaMa 70b and Mixtral 8x7B while keeping \(N_{\text{TFIDF}}\) and \(N_{\text{embed}}\) constant.

\begin{table*}[htbp]
\begin{tabularx}{\textwidth}{Xllllll}
\hline
\textbf{Model} & \textbf{\(N_{\text{TFIDF}}\)} & \textbf{\(N_{\text{embed}}\)} & \textbf{\texttt{UseDict}} & \textbf{BLEU} & \textbf{ChrF} & \textbf{ChrF++} \\ \hline
gpt-4-turbo & 0 & 0 & TRUE & 3 & 30.7 & 27.9 \\
gpt-4-turbo & 0 & 0 & FALSE & 0 & 30.8 & 26.9 \\
gpt-4-turbo & 10 & 0 & TRUE & 4 & 36.9 & 33.8 \\
gpt-4-turbo & 10 & 0 & FALSE & 0 & 33.4 & 29.9 \\
gpt-4-turbo & 0 & 10 & TRUE & 3.4 & 34.5 & 31.6 \\
gpt-4-turbo & 0 & 10 & FALSE & 0 & 31.4 & 27.8 \\
gpt-4-turbo & 5 & 5 & TRUE & \textbf{4.4} & \textbf{35.9} & \textbf{33} \\
gpt-4-turbo & 5 & 5 & FALSE & 0 & 33.7 & 29.9 \\
Mixtral 8x7B & 5 & 5 & TRUE & 3.5 & 26.8 & 24.6 \\
LlaMa 70b & 5 & 5 & TRUE & 0 & 27.7 & 24.7 \\
\end{tabularx}
\caption{Experiment results for the minicorpus of translations collected from a native Mambai speaker. \(N_{\text{TFIDF}}\) and \(N_{\text{embed}}\) represent the number of sentence pairs retrieved through TF-IDF and semantic embeddings, respectively. \texttt{UseDict} indicates whether dictionary entries are included in the prompt. While different hyperparameter combinations were tested for all models, we only report on the best configuration for the less performant models (Mistral 8x7B and LlaMa 70b).}
\label{tab:experiment_results_leo}
\end{table*}

\subsection{Error analysis}

We find that the large gap in performance across test sets is mostly due to differences in translation output, rather than differences in the source English text (Table~\ref{tab:tfidf_cosine_similarity}):
\begin{enumerate}
    \item Using TF-IDF representations of English sentences, we computed the cosine similarity  in the whole training set and the two tests sets, resulting in 0.021 for the manual test set and 0.017 for the native speaker test set, a relatively small difference. For the Mambai target reference, however, we get a 0.027 and 0.012 for the manual and native speaker's test sets, respectively, a much larger difference.
    \item LASER Semantic similarity between each test set and the training set are roughly equivalent at 0.42 and 0.40 for the manual and native speaker's test sets, respectively, on the English source side.
\end{enumerate}

\begin{table}[htbp]
\centering
\begin{tabular}{lccc}
\hline
\textbf{Similarity} & \textbf{Lang} & \textbf{Method} & \textbf{Score} \\
\hline
ManualTest x Train & eng & TF-IDF & 0.021 \\
NativeTest x Train & eng & TF-IDF & 0.017 \\
ManualTest x Train & mgm & TF-IDF & 0.027 \\
NativeTest x Train & mgm & TF-IDF & 0.012 \\
ManualTest x Train & eng & Semantic & 0.42 \\
NativeTest x Train & eng & Semantic & 0.40 \\
\hline
\end{tabular}
\caption{Similarity scores using TF-IDF cosine similarity and LASER semantic cosine similarity between the two test sets and the training set for English (source, eng) and Mambai (target, mgm) sentences.}
\label{tab:tfidf_cosine_similarity}
\end{table}

Through manual review of the translation differences in both test sets, we further identify the following potential causes for the large discrepancy in translation quality metrics:

\textbf{(1) Literal vs figurative translation}: As sentences in the language manual are made for learning, they tend to use more literal translations, which correspond to what LLMs produce. On the other hand, our test set translated by a native speaker often uses more idiosyncratic translation, further away from words used in from the source input.

\textbf{(2) Language variation}: The Mambai language has changed since 2001, when the Mambai Language Manual was published. In particular, we noted more usage of Portuguese and Tetun Dili words in our test set reference sentences, which might indicate that Mambai speakers mix more Tetun Dili and Portuguese in their Mambai since the two languages were chosen as official in the 2002 Constitution \cite{Constitution}.

\textbf{(3) Spelling}: Despite trying to stay close to spelling used in the Mambai Language Manual, we found that our test set at times uses different spelling than the language manual (e.g. less hyphenation, some letters missing). This reinforces our view that oral languages like Mambai are better covered by speech datasets.

\section{Related Work}

Traditionally, neural MT systems are trained on parallel corpora of aligned sentence pairs \cite{duong2017natural}. Low-resource languages tend to have orders of magnitude less sentences available than higher-resource languages \cite{arivazhagan2019massively}. To compensate for this lack of data, previous research found that low-resource MT accuracy can be improved through leveraging multilingual translation models that include better-resourced but related languages \cite{arivazhagan2019massively, fan2020englishcentric, nllb2022}. Other techniques include pre-training on monolingual data \cite{lample2018unsupervised}, the incorporation of audio data that shares an embedding space with text data \cite{communication2023seamlessm4t}, and the generation of synthetic parallel sentences \cite{edunov2018understanding}, including by leveraging bilingual dictionaries \cite{duan2020bilingual}.

In parallel, large language models have shown increased ability to translate, at times surpassing specialised encoder-decoder MT systems \cite{robinson-etal-2023-chatgpt}. Finding the right prompt recipe for increased MT accuracy using LLMs has been a topic of research \cite{pmlr-v202-zhang23m, li-etal-2022-prompt}, with findings that few-shot prompting often improves MT accuracy \cite{pmlr-v202-zhang23m}, and that the type of sentences used as few-shot examples can have a large influence on accuracy \cite{moslem2023adaptive}. Dynamic adaptation of the prompt by retrieving example sentences that are close to the input text \cite{kumar2023ctqscorer}, or dictionary entries for words that appear in the source \cite{ghazvininejad2023dictionarybased} can further improve MT accuracy.

The applicability of common LLM prompting techniques when translating into very low-resource languages is unclear, given these languages might not be represented at all during LLM pretraining. \citet{tanzer2024mtob} partially addresses this issue by focusing on MT between English and Kalamang, an endangered Papuan language, using a single grammar book. Experimenting with different models (Claude 2, LlaMa, gpt-3.5, gpt-4), and different prompt setups (injecting sentences close to the input, dictionary entries, and the grammar explanations found in the book), they achieve up to 45.8 ChrF on the English to Kalamang direction. However, they work with a test set that is a random subset of sentences found in the book, raising issues around the applicability of their results to text translated by a different author, or to domains not covered in the grammar book.

Recognising the potential of LLMs for MT, and the importance of in-context examples used in prompting, our work experiments with retrieval-augmented LLM prompting for translation into a low-resource language. We test translation quality on both a subset of sentences coming from the language manual used as corpus, and a test set specially translated by a native Mambai speaker for this project.

\section*{Conclusion}

In this paper, we introduced a novel corpus for the Mambai language, a language with around 200,000 native speakers that had virtually no NLP resources. Our corpus includes bilingual dictionaries in both directions for English-Mambai, a set of 1,187 parallel sentences from a language manual published in 2001, and a set of 50 parallel sentences translated by a native Mambai speaker. Our experiments on few-shot LLM prompting for English to Mambai translation showed that moderate MT quality can be achieved for test sentences very close to the original corpus, but MT quality decreases significantly for sentences that come from a separate corpus, thus highlighting the need for using test sets that do not come from the same material as original examples used in prompting. We think LLMs offer a flexible approach for integrating scarce resources in different formats (dictionary entries, parallel sentences), and few-shot prompting shows potential in improving low-resource MT using general purpose LLMs.

\section*{Limitations}

The sentences used in both training set (from the Mambai Language Manual) and test sets tend to be rather short and simple, which raises questions around translation quality for longer sentences, or for technical domains that get little coverage in our corpus (e.g. health or legal text).

Mambai has no standard orthography. Even though the native Mambai speaker we collaborated with tried to follow spelling close to that used in the language manual, we expect that variances in spelling still negatively impacted the test BLEU score. This stresses the need for heightened focus on audio for primarily spoken languages like Mambai \cite{chrupala-2023-putting}.

While we were able to gather a test set from a native Mambai speaker, they did not evaluate translation quality for MT-translated text; instead we relied solely on automated MT metrics. While BLEU tends to be a reliable measure of MT quality for morphologically simple languages like Mambai \cite{bleu}, we would have preferred to dig deeper into the shortcomings of our LLM-generated translations.

Lastly, Mambai has a simple grammar and morphology, which might make it particularly prone to MT quality improvement using few-shot prompting. Therefore, our results might not translate well on more morphologically complex languages.

\section*{Future Work}

This work focused solely on Mambai, without leveraging resources from related languages that have more resources, such as Tetun Dili, Portuguese, or Indonesian. In future work, we would like to investigate the addition of Tetun Dili sentences to the prompt, especially for domain-specific text that might be very poorly covered by our small Mambai corpus, but that could be covered by a larger Tetun Dili corpus.

In terms of finding the right recipe for prompting, future endeavours could use a more systematic approach, similar to \citet{kumar2023ctqscorer} which uses a regression model for example selection. Additionally, more retrieval techniques could be tested, e.g. bag of words, or even ChrF similarity between the input and English source side.

In this paper, we used general purpose LLMs that likely saw little to no Mambai text during pretraining. We think future work could experiment with continuous pretraining on Mambai, or languages related to Mambai, before prompting, similar to approaches in \citet{xu2024contrastive-alma} and \citet{alves2024tower}.

\section*{Acknowledgements}

We thank Pr. Geoffrey Hull (Macquarie University), author of the Mambai Language Manual, for authorising usage of his work as part of this research. Pr. Hull remains the holder of copyright protecting this intellectual property.

\section*{References}
\label{sec:reference}

\bibliographystyle{lrec-coling2024-natbib}
\bibliography{paper}

\section*{Language Resource References}
\label{lr:ref}
\bibliographystylelanguageresource{lrec-coling2024-natbib}
\bibliographylanguageresource{languageresource}

\end{document}

%% file: data_extraction_process.tex
\usetikzlibrary{shapes.geometric, arrows.meta, positioning}

\tikzset{
    base/.style = {draw, text width=2.5cm, align=center, minimum height=1em, font=\small},
    inout/.style = {base, rectangle, rounded corners},
    process/.style = {base, rectangle, fill=blue!30},
    gate/.style = {base, chamfered rectangle, fill=red!30},
}

\begin{tikzpicture}[node distance=1.2cm and 0.5cm, auto]
    \node [process] (scanocr) {Scan \& OCR};
    \node [inout, right=1.5cm of scanocr] (allword) {Word document};
    \node [inout, above=1.5cm of allword] (worddicts) {Word files containing dictionaries};
    \node [inout, below=1.5cm of allword] (wordsentences) {Word file containing parallel sentences};
    \node [process, right=1.5cm of worddicts] (extractdict) {Extract dictionaries in json format using regex};
    \node [process, right=1.5cm of wordsentences] (rawsentences) {Extract Mambai and English sentences in separate files using font weight hints};
    \node [gate, right=5cm of allword] (hunalign) {Hunalign: align sentences};
    \node [inout, right=5.7cm of extractdict] (dictoutput) {Dictionaries for English-Mambai and Mambai-English};
    \node [inout, right=2cm of hunalign] (para) {Parallel aligned English-Mambai sentences};

    \draw[-Latex] (scanocr) -- (allword);
    \draw[-Latex] (allword) -- (worddicts);
    \draw[-Latex] (allword) -- (wordsentences);
    \draw[-Latex] (worddicts) -- (extractdict);
    \draw[-Latex] (wordsentences) -- (rawsentences);
    \draw[-Latex] (extractdict) -| (hunalign);
    \draw[-Latex] (rawsentences) -| (hunalign);
    \draw[-Latex] (hunalign) -- (para);
    \draw[-latex] (extractdict) -- (dictoutput);
\end{tikzpicture}

%% file: translation_process.tex
\usetikzlibrary{shapes.geometric, arrows.meta, positioning}

\tikzset{
    base/.style = {draw, text width=2.5cm, align=center, minimum height=1em, font=\small},
    inout/.style = {base, rectangle, rounded corners},
    process/.style = {base, rectangle, fill=blue!30},
    gate/.style = {base, chamfered rectangle, fill=red!30},
}

\begin{tikzpicture}[node distance=1.2cm and 0.5cm, auto]
    \node [inout] (input1) {Example sentences from manual};
    \node [inout, below=of input1] (input2) {Input: English sentence from test set};
    \node [inout, below=of input2] (input3) {Dictionary entries from manual};
    \node [process, right=1.5cm of input1] (process1) {Selection of example sentences where source is close to input};
    \node [process, right=1.5cm of input3] (process2) {Filter dictionary entries for words that appear in input sentences};
    \node [process, right=5cm of input2] (process3) {Prompt construction from example sentences and dictionary entries};
    \node [gate, right=2cm of process3] (llm) {LLM};
    \node [inout, right=2cm of llm] (output) {Output: translated sentence in Mambai};

    \draw[-latex] (input1) -- (process1);
    \draw[-latex] (input2) -| (process1);
    \draw[-latex] (input2) -| (process2);
    \draw[-latex] (input3) -- (process2);
    \draw[-latex] (process1) -| (process3);
    \draw[-latex] (process2) -| (process3);
    \draw[-latex] (process3) -- (llm);
    \draw[-latex] (llm) -- (output);

\end{tikzpicture}